\documentclass[letterpaper, 10 pt, conference]{ieeeconf}
\IEEEoverridecommandlockouts    %
\usepackage{hyperref}
\usepackage{threeparttable}
\usepackage{algorithm}
\usepackage{cite}
\usepackage{pifont}
\usepackage{ragged2e}

\hypersetup{
    colorlinks=true,
    linkcolor=magenta,
    filecolor=magenta,      
    urlcolor=magenta,
}

\usepackage{wrapfig}
\usepackage{lipsum}
\usepackage{xspace}
\newcommand{\etal}{\emph{et al.}\xspace}
\usepackage{multirow}
\usepackage[table]{xcolor} %
\definecolor{lightgray}{RGB}{220, 220, 220} %
\usepackage{caption}
\usepackage{subcaption}

\usepackage{graphics}           
\usepackage{times}              
\usepackage{amsmath}            
\usepackage{amssymb}            
\usepackage{graphicx}
\usepackage{algorithm}
\usepackage[noend]{algpseudocode}
\usepackage{booktabs}
\usepackage{color}
\definecolor{instructioncolor}{rgb}{.5,.5,.5}

\usepackage[font=small]{caption}

\def\eqref#1{Eq.~(\ref{#1})}

\makeatletter
\usepackage{xspace}
\DeclareRobustCommand\onedot{\futurelet\@let@token\@onedot}
\def\@onedot{\ifx\@let@token.\else.\null\fi\xspace}

\def\etal{{et al}\onedot}
\makeatother

\usepackage{array}
\newcolumntype{L}[1]{>{\raggedright\let\newline\\\arraybackslash\hspace{0pt}}m{#1}}
\newcolumntype{C}[1]{>{\centering\let\newline\\\arraybackslash\hspace{0pt}}m{#1}}
\newcolumntype{R}[1]{>{\raggedleft\let\newline\\\arraybackslash\hspace{0pt}}m{#1}}

\usepackage{bm}
\usepackage{multirow}
\usepackage{rotating}  %

\renewcommand{\thefigure}{\Roman{figure}}
\newcommand{\myblue}[1]{{\color{blue}#1}}
\definecolor{mygreen_rgb}{RGB}{0,0,0}

\title{\LARGE \bf Zero-Shot Temporal Interaction Localization for Egocentric Videos}

\author{Erhang~Zhang$^{*}$, Junyi~Ma$^{*}$, Yin-Dong Zheng, Yixuan Zhou, Hesheng~Wang$^{\dag}$
\thanks{Erhang~Zhang, Junyi~Ma, Yin-Dong~Zheng, Yixuan Zhou, and Hesheng~Wang are with IRMV Lab, the Department of Automation, Shanghai Jiao Tong University.}
\thanks{$^{*}$Equal contribution}
\thanks{$^{\dag}$Corresponding author email: wanghesheng@sjtu.edu.cn}
}

\begin{document}
\maketitle

\IEEEpeerreviewmaketitle
\thispagestyle{empty}
\pagestyle{empty}

\begin{abstract}
Locating human-object interaction (HOI) actions within video serves as the foundation for multiple downstream tasks, such as human behavior analysis and human-robot skill transfer. Current temporal action localization methods typically rely on annotated action and object categories of interactions for optimization, which leads to domain bias and low deployment efficiency. Although some recent works have achieved zero-shot temporal action localization (ZS-TAL) with large vision-language models (VLMs), their coarse-grained estimations and open-loop pipelines hinder further performance improvements for temporal interaction localization (TIL). To address these issues, we propose a novel zero-shot TIL approach dubbed \textit{EgoLoc} to locate the timings of grasp actions for human-object interaction in egocentric videos.
EgoLoc introduces a self-adaptive sampling strategy to generate reasonable visual prompts for VLM reasoning. By absorbing both 2D and 3D observations, it directly samples high-quality initial guesses around the possible contact/separation timestamps of HOI according to 3D hand velocities, leading to high inference accuracy and efficiency. In addition, EgoLoc generates closed-loop feedback from visual and dynamic cues to further refine the localization results. Comprehensive experiments on the publicly available dataset and our newly proposed benchmark demonstrate that EgoLoc achieves better temporal interaction localization for egocentric videos compared to state-of-the-art baselines. We will release our code and relevant data as open-source at \url{https://github.com/IRMVLab/EgoLoc}. 
\end{abstract}

\section{Introduction}
\label{sec:intro}

\begin{figure}
  \centering
  \includegraphics[width=1\linewidth]{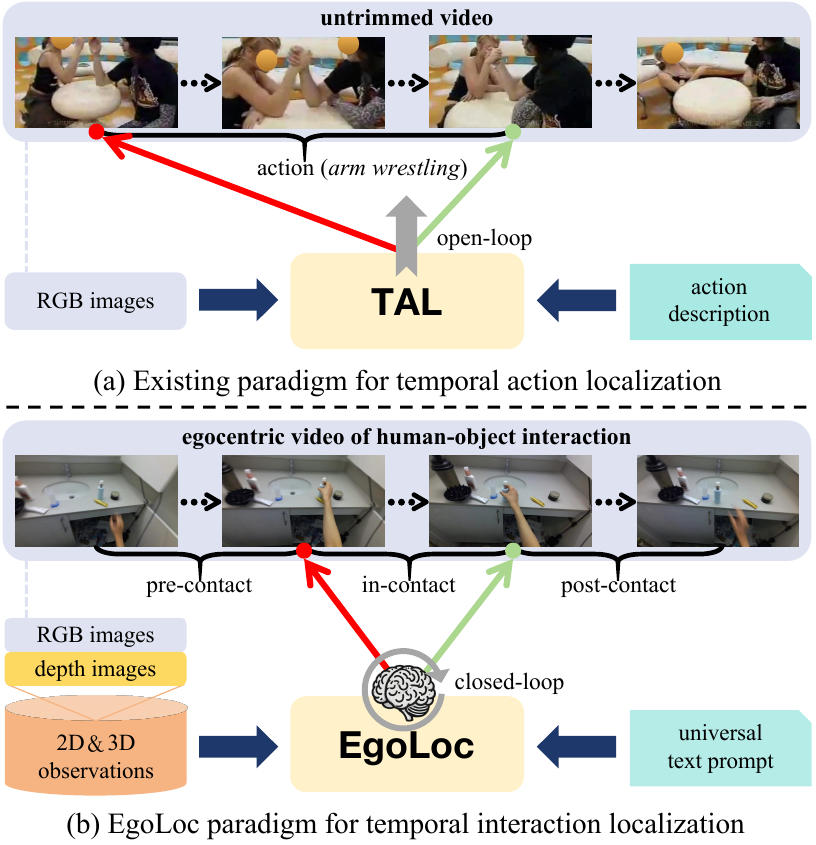}
  \caption{Compared to the TAL paradigm (a) that localizes a specific action in an untrimmed video with RGB images and action descriptions, TIL aims to find the frames closest to the start and end of a human-object in-contact period. Our proposed EgoLoc paradigm (b) absorbs both 2D and 3D observations and implements accurate temporal interaction localization with a universal text prompt in a closed-loop form.}
  \label{fig:motivation_maintext}
  \vspace{-0.7cm}
\end{figure}

Temporal action localization (TAL)~\cite{shou2016temporal,gao2023pami,phan2024zeetad,li2024detal} has long been a focal issue in the computer vision community. It aims to determine the time durations of specific actions from a long untrimmed video as shown in Fig.~\ref{fig:motivation_maintext}(a), and provides prior knowledge for motion analysis and prediction~\cite{ma2024diff,wang2023temporal}. In this work, we step further toward temporal interaction localization (TIL) in the context of TAL, considering its possible applications to complex downstream tasks such as human behavior analysis and human-robot skill transfer. As shown in Fig.~\ref{fig:motivation_maintext}(b), we regard any human-object interaction (HOI) process as three stages, including the pre-contact, in-contact, and post-contact stages. Therefore, TIL attempts to locate the egocentric frames closest to the start and end of a human-object in-contact period, accurately identifying the three stages. The fundamental distinction between TIL and TAL lies in the granularity of temporal localization, particularly in the emphasis on accurately capturing the precise timings of hand-object in-contact states.

While existing TAL methods have achieved promising results in localizing actions using verb-noun forms, they face limitations in the following four aspects when adapted to TIL tasks in egocentric videos. (1) \textit{Granularity}: The existing TAL methods are primarily developed to find the action/task transition regardless of hand-object contact states. Therefore, they cannot precisely locate the fine-grained timestamps at which a human hand makes contact with and separates from a target object in TIL tasks. (2) \textit{Generalization capability}: Human-object interaction involves various action/object categories. However, most TAL methods require labor-intensive annotations for model optimization, resulting in low deployment efficiency and poor generalization ability. Although some zero-shot temporal action localization (ZS-TAL) methods~\cite{wake2024open,gupta2024open} have been proposed for better generalizability, their performance is still limited by predefined action/object categories in the text prompts. (3) \textit{Perception capability}: 
TAL methods typically rely on 2D RGB images as input. They overlook human motion in 3D space which could provide richer interaction information with an absolute scale to enhance the understanding of HOI.
(4) \textit{Closed-loop capability}: TAL methods are generally designed as open-loop frameworks in an end-to-end manner, resulting in high estimation uncertainties. A more suitable framework should introduce a closed-loop design, where the model automatically assesses its localization results, provides feedback, and updates its output to reduce uncertainties and improve accuracy.

To this end, we propose a novel zero-shot temporal interaction localization (ZS-TIL) paradigm named \textit{EgoLoc} for egocentric videos. It locates the timestamps at which a human hand makes contact with and separates from an object. We attend to grasp actions, the most frequent action in human-object interaction. EgoLoc first combines 2D RGB images and 3D point clouds to extract 3D hand velocities. Then, a specially designed self-adaptive sampling strategy generates the high-quality initial guesses of the frames closest to the start (contact) and end (separation) of interaction from the frames with the lowest hand velocities. 
This follows the fact that hands always maintain relatively low speeds when they come into contact with or separate from a target object. Notably, we extract hand velocities in the 3D global coordinate system to avoid depth ambiguity and scale aliasing from 2D observations. Subsequently, with a universal text prompt, a large vision-language model (VLM) directly locates contact and separation timestamps within the initial guesses. Moreover, we coordinate the output for self-checking with closed-loop feedback from visual and dynamic cues, refining localization results. This reduces localization uncertainties and further improves accuracy. Compared to existing ZS-TAL paradigms, EgoLoc samples high-quality keyframes of interest and implements temporal localization more accurately with a closed-loop form (see Fig.~\ref{fig:motivation_maintext}). The main contributions of this work are as follows: 

\begin{itemize}
    \item We propose a novel paradigm namely EgoLoc for zero-shot temporal interaction localization in egocentric videos. It is developed based on the off-the-shelf VLM and performs interaction localization with the universal text prompt, eliminating the need for specific action-object descriptions during training and inference.
    \item To our best knowledge, EgoLoc is the first work to integrate 3D hand velocities estimated from 2D and 3D perception into temporal localization. We design a new self-adaptive sampling strategy to generate high-quality initial guesses of the contact and separation timestamps. 
    \item EgoLoc introduces a closed-loop scheme in the TIL reasoning process, refining the results through feedback from visual and dynamic cues to reduce estimation uncertainties and improve localization accuracy.
    \item Extensive experiments on the publicly available dataset and our newly proposed benchmark demonstrate that EgoLoc significantly outperforms existing approaches in the task of temporal interaction localization.
\end{itemize}

\begin{figure*}
  \centering
  \includegraphics[width=1\linewidth]{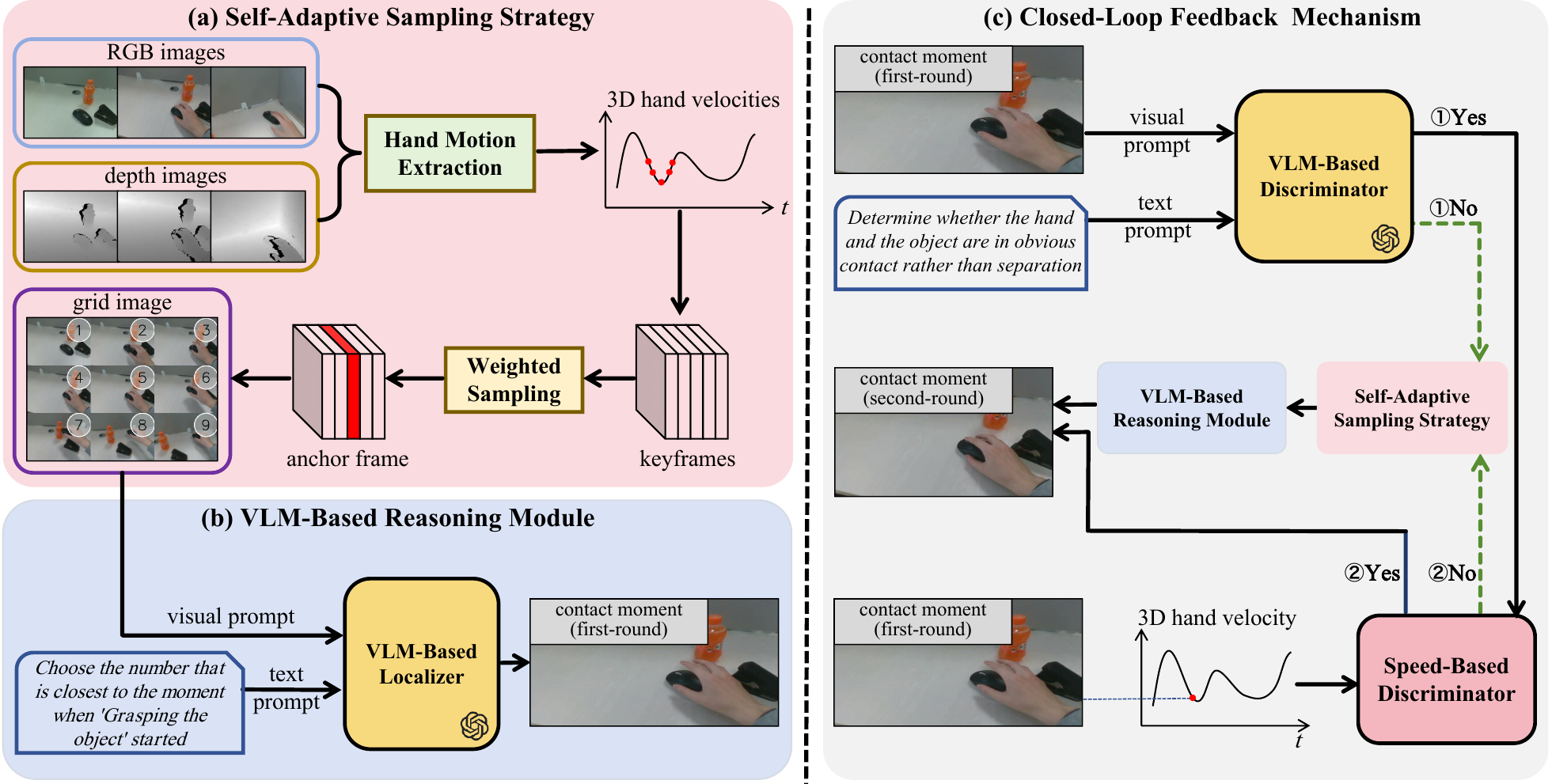}
  \caption{\textbf{The overall pipeline of  EgoLoc}: EgoLoc integrates the self-adaptive sampling strategy, the VLM-based reasoning module, and the closed-loop feedback mechanism. The estimation of the contact timestamp is illustrated as an example for clearer clarification. Firstly, the self-adaptive sampling strategy samples the initial guesses of the contact timestamp using 3D hand velocities from RGB and depth images. Then, the VLM-based reasoning module generates the first-round contact timestamp based on the visual prompt. Finally, the closed-loop feedback mechanism outputs the refined second-round timestamp by assessing hand-object contact states and velocities.}
  \label{fig:overall}
  \vspace{-0.5cm}
\end{figure*}

\vspace{-0.1cm}
\section{Related work}
\label{sec:related_work}

\subsection{VLM-Based Temporal Action Localization}
Temporal action localization (TAL)~\cite{shou2016temporal,gao2023pami,phan2024zeetad,li2024detal,yan2023unloc} has been widely explored in recent years. Unlike traditional action recognition techniques~\cite{zheng2020dynamic,bao2021evidential,Shiota_2024_WACV,peng2024referring,abdullah2025ual}, TAL estimates the start and end times of specific actions. With the rise of large vision-language models, many TAL methods now leverage pre-trained VLMs, which offer powerful vision-language features and zero-shot inference capabilities for improved action localization. As a pioneering work, Ju \etal~\cite{ju2022prompting} efficiently adapt CLIP~\cite{radford2021learning} to various video understanding tasks by optimizing task-specific prompt vectors and temporal Transformer. Based on this work, UnLoc~\cite{Yan_2023_ICCV} also exploits VLM to extract image-language features, which are processed by the stacked Transformers and a feature pyramid to predict a frame-level relevancy score and start/end displacements. STALE~\cite{nag2022zero} follows CLIP style classification and further introduces class-agnostic representation masking, achieving generalization to unseen classes. Different from STALE, ZEETAD~\cite{phan2024zeetad} correlates semantic features from VLM with a dynamic foreground mask. More recently, T3AL~\cite{liberatori2024test} achieves zero-shot TAL by adapting the VLM at test time with self-supervision. In these works, VLMs are typically used as a semantic feature extractor for input images. In contrast, some other works adopt the strong reasoning ability of large language models (LLMs) and VLMs to achieve zero-shot and open-vocabulary TAL. For example, Ju \etal~\cite{ju2023multi} propose decomposing actions by prompting
LLMs for various action attributes, improving the discriminative power of constructed classifiers. In OVTAL~\cite{gupta2024open}, text encoding is conducted on the class-specific language descriptions generated by LLM. Recently, T-PIVOT~\cite{wake2024open} directly exploits OpenAI's GPT-4o to reason about the timings of specific actions from tiled images sampled from a long video. It follows PIVOT~\cite{nasiriany2024pivot} to iteratively locate the frames of interest with gradually decreasing window sizes. In this work, we also develop EgoLoc based on the off-the-shelf VLM. Unlike T-PIVOT which attends to coarse action transitions, EgoLoc extracts more fine-grained timings of human-object contact states in the task of temporal interaction localization for egocentric videos. Additionally, we propose using additional 3D perception data to enhance human behavior understanding, in contrast to those TAL methods that rely solely on 2D images/videos. We also introduce a closed-loop scheme to further improve the VLM-based localization performance.

\subsection{VLM-Enhanced Egocentric Vision}

Egocentric vision~\cite{plizzari2024outlook,betancourt2015evolution} refers to capturing and analyzing visual data from first-person videos, typically recorded by head-mounted camera devices such as smart glasses. Some egocentric vision techniques like action recognition/anticipation~\cite{zou2025towards,zhou2023can,furnari2020rolling,mascaro2023intention} and HOI prediction~\cite{mur2024aff,ma2024diff,liu2022joint} have made significant progress in the past decades. In recent years, extensive works have adopted advanced VLMs to enhance reasoning soundness and generalizability. For example, Mittal~\etal~\cite{mittal2024can} combine Q-former~\cite{li2023blip} and LLM to build a large video-language model for action anticipation. Rai \etal~\cite{rai2024strategies} achieve weakly supervised affordance grounding with the knowledge from VLMs. Egothink~\cite{cheng2024egothink} presents the first-person perspective reasoning ability of multiple advanced VLMs. Compared to these egocentric vision techniques, our proposed EgoLoc is a novel VLM-based paradigm for egocentric temporal interaction localization, expanding the scope of egocentric vision research.

\section{Proposed Method}
\label{sec:proposed_method}

\subsection{Task Definition}
\label{sec:task_def}
Temporal interaction localization (TIL) can be regarded as a more fine-grained task stemming from temporal action localization.
Here we first detail the definition of the TIL task.
Given an egocentric video with $N$ sequential RGB images $\mathcal{I}=\{I_t\}_{t=1}^{N} (I_t \in \mathbb{R}^{3\times h\times w})$ and depth images $\mathcal{D}=\{D_t\}_{t=1}^{N} (D_t \in \mathbb{R}^{1\times h\times w})$, TIL models aim to estimate the contact timestamp $T_\text{c}$ and separation timestamp $T_\text{s}$, which denote the timings at which a human hand makes contact with and separates from a target object, respectively. As mentioned in Sec.~\ref{sec:intro}, each egocentric video clip of human-object interaction can be split into the pre-contact, in-contact, and post-contact stages by $T_\text{c}$ and $T_\text{s}$. Therefore, TIL models have the potential to trim out the video segments of interest in the interaction for the following downstream tasks.

\subsection{Overall Pipeline of EgoLoc}
\label{sec:so}

The overall TIL pipeline of our proposed EgoLoc is depicted in Fig.~\ref{fig:overall}, taking the hand-object contact timestamp estimation as an example. We first design a self-adaptive sampling strategy shown in Fig.~\ref{fig:overall}(a) to select the initial guesses of the contact timestamp. The hand motion extraction module extracts 3D hand velocities for each frame from sequential RGB images and depth images. Then, we adaptively sample an anchor frame from those frames with the lowest hand velocities in the first half of the video. A grid image is further constructed by tiling the multiple frames adjacent to the anchor frame, which are annotated in chronological order automatically. The anchor frame and its adjacent frames are regarded as high-quality initial guesses for the contact timestamp. Our self-adaptive sampling strategy is motivated by the fact that the hand moves more slowly at the timestamps of contact and separation, while moving faster during the pre-contact, in-contact, and post-contact stages.

In the following VLM-based reasoning module shown in Fig.~\ref{fig:overall}(b), we regard the grid image as the visual prompt and build a universal text prompt for the VLM-based localizer, which directly estimates the first-round contact timestamp. The one-step inference process of the VLM-based localizer significantly improves the efficiency of the state-of-the-art (SOTA) ZS-TAL method~\cite{wake2024open} with multiple iteration steps, while maintaining high estimation accuracy. This mainly benefits from the high-quality design of the visual prompt.

To further refine the contact timestamp identified by the VLM-based localizer, we propose a closed-loop feedback mechanism as shown in Fig.~\ref{fig:overall}(c). 
The combination of the VLM-based discriminator and speed-based discriminator helps to judge whether the hand and the object at the estimated contact timestamp are visually in obvious contact with a relatively low hand velocity. Then, we can decide whether to re-localize from the self-adaptive sampling in Fig.~\ref{fig:overall}(a) with additional constraints, closing the loop of the TIL process. Our feedback mechanism ultimately outputs the refined second-round timestamp estimation.

After we determine the contact timestamp, we extract the video segment after the estimated contact timestamp as the input for EgoLoc to further find the separation timestamp. The overall localization process is similar to the above-mentioned contact timestamp estimation. We ultimately achieve temporal interaction localization after the contact and separation timestamps are both determined.

\begin{figure}[t]
  \centering
    \includegraphics[width=1\linewidth]{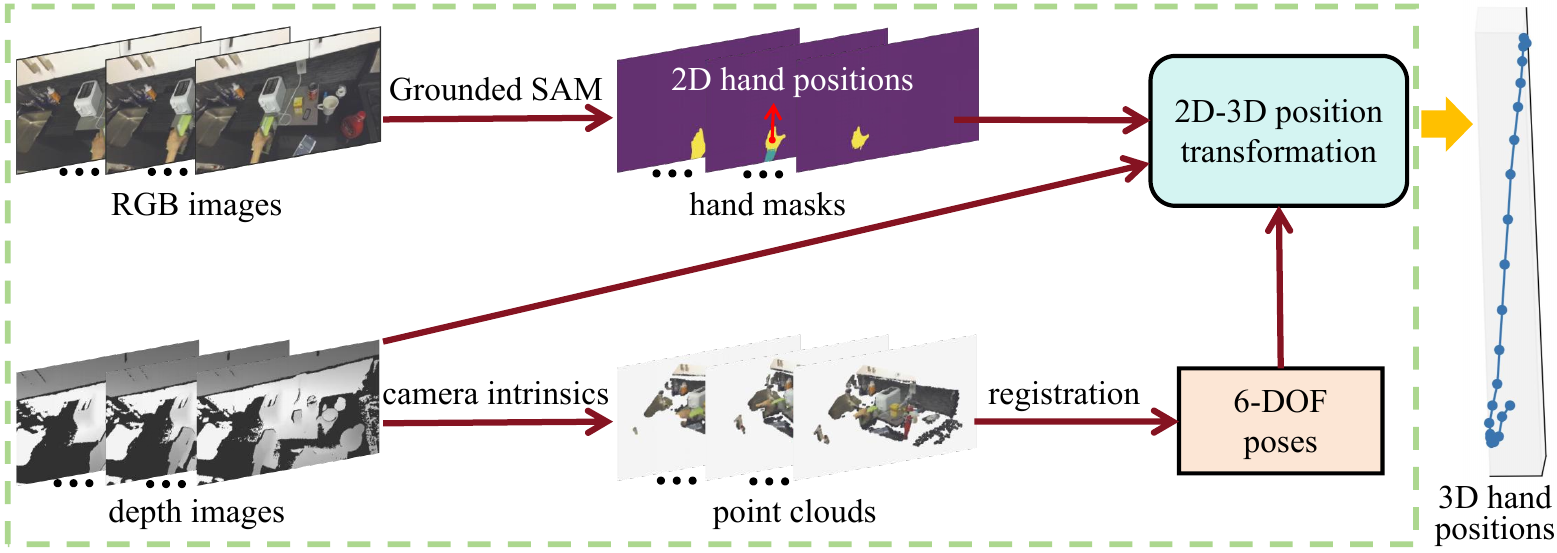}
  \caption{The hand motion extraction module generates 2D hand masks with Grounded SAM~\cite{ren2024grounded} and transforms the hand depth in each camera coordinate system to 3D hand positions in the global coordinate system. The 3D hand positions are further used to calculate 3D hand velocities.}
  \label{fig:hand_ext}
  \vspace{-0.7cm}
\end{figure}

\subsection{Self-Adaptive Sampling}
\label{sec:sas}

EgoLoc is developed based on the off-the-shelf VLM, which is expected to locate hand-object contact and separation timestamps with multiple egocentric images as the visual prompt. In this work, we require the VLM-based localizer in Fig.~\ref{fig:overall}(b) to select the frames closest to the moments of contact and separation from the visual prompt. Therefore, the high quality of the visual prompt is essential, since the sampled images within it define the approximate localization scope affecting the rationality of the final results. The previous work~\cite{wake2024open} uniformly samples images from the entire video for the initial visual prompt, which leads to inaccurate localization results from VLM and low inference efficiency. In contrast, in this work, we propose a self-adaptive sampling strategy based on hand velocity analysis, to produce high-quality initial guesses of contact/separation timestamps as the visual prompt for the VLM-based localizer. The self-adaptive sampling strategy incorporates both 2D and 3D egocentric observations, enhancing the perception capability for 3D hand motion. It exploits the fact that a hand typically has a relatively lower velocity when it makes contact with or separates from the target object. Here we do not use 2D hand motion in the image plane, considering depth ambiguity and scale aliasing from 2D observations.

Specifically, we first capture 3D hand velocities through the hand motion extraction module. As Fig.~\ref{fig:hand_ext} shows, we exploit Grounded SAM~\cite{ren2024grounded} prompted with \textit{hand} to obtain hand masks on the sequential images $\mathcal{I}$. By averaging the masked pixel coordinates in each image $I_t \in\mathcal{I}$, we obtain 2D hand positions $\mathcal{H}^{\text{2D}}=\{H^{\text{2D}}_t\}_{t=1}^N (H^{\text{2D}}_t \in \mathbb{R}^{2})$. Then, with the camera intrinsics and the sequential depth images $\mathcal{D}$, we calculate 3D hand positions $\mathcal{H}^{\text{3D,cam}}=\{H_t^{\text{3D,cam}}\}_{t=1}^N (H_t^{\text{3D,cam}} \in \mathbb{R}^{3})$ in the camera coordinate systems of all timestamps. Next, we perform registration between sequential 3D point clouds from $\mathcal{D}$, obtaining the transformation matrix $\mathcal{M}=\{M_t\}_{t=1}^N (M_t \in \mathbb{R}^{4 \times 4})$. $M_t$ denotes the relative 6-DOF pose between the $t$\,th frame and the first frame $(t=1)$. $\mathcal{M}$ transform $\mathcal{H}^{\text{3D,cam}}$ to $\mathcal{H}^{\text{3D,glob}}=\{H_t^{\text{3D,glob}}\}_{t=1}^N (H_t^{\text{3D,glob}} \in \mathbb{R}^{3})$ in the camera coordinate system of the first frame, which is regarded as the global coordinate system in this work. 

Then we use $\mathcal{H}^{\text{3D,glob}}$ to calculate the 3D hand velocities $\mathcal{V} = \{V_t\}_{t=1}^N (V_t \in \mathbb{R}^{1})$ for each timestamp of the egocentric video by $V_t=(H_{t+1}^{\text{3D,glob}}-H_{t}^{\text{3D,glob}})/\delta$, where $\delta$ is the fixed time interval between adjacent frames. We simply set $V_N=V_{N-1}$ since there is no $H^{\text{3D,glob}}_{N+1}$ for $V_N$. $N_\text{key}$ frames with the lowest velocities are further selected to compose the set of keyframes $\mathcal{I}_\text{key}$. For the contact moment, we select $\mathcal{I}_\text{key}$ from the first half of the video, while for the separation moment, $\mathcal{I}_\text{key}$ is selected from the segment after the ultimate estimated contact timestamp. Then, we perform uniformly sampling on the keyframes $\mathcal{I}_\text{key}$, obtaining one anchor frame $I_\text{init} \in \mathcal{I}_\text{key}$ as one of the initial high-quality guess when the hand makes contact with or separates from an object. 

Moreover, we sample $N_\text{adj}^2-1$ adjacent frames associated with the symmetry of $I_\text{init}$ to get the enriched initial guesses $\mathcal{I}_\text{adj}$. We attend to $\mathcal{I}_\text{adj}$ for the following VLM reasoning because it is highly likely that the optimal contact moment is located at or near $I_\text{init}$. Besides, these consecutive frames can coherently capture the specific state transition between the pre-/post-contact and in-contact stages, compared to $\mathcal{I}_\text{key}$ which may be temporally discrete. Following~\cite{wake2024open}, we then construct a grid image $G \in \mathbb{R}^{N_\text{adj}h\times N_\text{adj}w}$ by tiling all the $N_\text{adj}^2$ frames encompassing $\mathcal{I}_\text{adj}$ and $I_\text{init}$, which are annotated in chronological order automatically. The bottom-left corner of Fig.~\ref{fig:overall}(a) illustrates the exampled grid image with $N_\text{adj}=3$, where $I_\text{init}$ holds the index number 5. $G$ is regarded as the output of our proposed self-adaptive sampling strategy.

\subsection{VLM-Based Reasoning}
\label{sec:vbr}

We exploit $G$ as the visual prompt and design a text prompt shown in Fig.~\ref{fig:overall}(b) for the following VLM-based localizer, to infer the timestamps of HOI contact and separation. The text prompt is universal for all HOI scenarios, maintaining generalizability. The VLM-based localizer uses an off-the-shelf VLM to directly output the index of the tiled images from $G$, to get the contact timestamp $\hat{T}_\text{c}$ and the separation timestamp $\hat{T}_\text{s}$. Note that the VLM-based localizer implements one-step inference rather than multiple inference iterations. This is because the visual prompt $G$ is elaborated by our self-adaptive sampling strategy offering high-quality initial guesses, compared to the coarsely sampled image prompts in the SOTA ZS-TAL method~\cite{wake2024open}. This significantly improves the localization efficiency.

\subsection{Closed-loop Feedback}
\label{sec:clf}
After VLM-based reasoning, we further propose a closed-loop feedback mechanism to refine the contact timestamp $\hat{T}_\text{c}$ and separation timestamp $\hat{T}_\text{s}$ estimated by the VLM-based localizer. Through closed-loop feedback, we enable EgoLoc to achieve self-improvement according to both visual cues and dynamic cues, resulting in more accurate TIL results.

Specifically, the feedback refers to the justification of whether self-adaptive sampling and VLM-based reasoning need to be re-implemented. It is generated by assessing the visual and dynamic cues of the first-round contact and separation moments. Firstly, we use the RGB image at the first-round contact or separation timestamp as a visual prompt, combined with a predefined text prompt shown in Fig.~\ref{fig:overall}(c). Both prompts are absorbed by the VLM-based discriminator, which uses the same VLM as the VLM-based localizer. It determines whether the estimated contact or separation moment is in the correct visual states. If the RGB images at $\hat{T}_\text{c}$ and $\hat{T}_\text{s}$ are judged by the VLM-based discriminator that the hand and the object are in a clearly contact state (\ding{172}Yes), the estimations are considered to have the correct visual states. Otherwise (\ding{172}No), the estimated timestamps are deemed to be wrong and we need to re-implement self-adaptive sampling as well as VLM-based reasoning, closing the loop of temporal interaction localization. This justification with visual cues follows the fact that at the moments of hand-object contact and separation, there should be no significant visual gap between the hand and object positions, which the VLM can capture. The estimated timestamps with correct visual states (\ding{172}Yes) are further checked by the speed-based
discriminator with dynamic cues. Referring to the previously extracted 3D hand velocities, we determine whether the velocities at the estimated timestamps fall within the slowest $\mu_\text{vel}\%$ of all velocities in the holistic input video clip. If yes (\ding{172}Yes$\rightarrow$\ding{173}Yes), the estimated contact or separation moment is accepted as the ultimate TIL output. Otherwise (\ding{172}Yes$\rightarrow$\ding{173}No), the estimated timestamps with high dynamics are deemed to be wrong, and we need to re-implement self-adaptive sampling and VLM-based reasoning. This justification with dynamic cues is motivated by the fact that the hand moves slowly at the timestamps of contact and separation, which has been mentioned in Sec.~\ref{sec:sas}. 

\vspace{-0.04cm}
Here we describe how to re-implement self-adaptive sampling with new constraints in the two cases (\ding{172}No) and (\ding{172}Yes$\rightarrow$\ding{173}No). Self-adaptive sampling triggered by the closed-loop feedback mechanism (green dashed lines in Fig.~\ref{fig:overall}(c)) differs from the counterpart of the first round in terms of the selection of keyframes $\mathcal{I}_\text{key}$. In the case (\ding{172}No) of the contact moment, the more accurate contact timestamp should be after the estimated $\hat{T}_\text{c}$, since (\ding{172}No) indicates that the hand and object are not yet in contact at $\hat{T}_\text{c}$. Therefore, we generate $\mathcal{I}_\text{key}$ by selecting $N_\text{key}$ frames with the lowest hand velocities from the video segment after $\hat{T}_\text{c}$. In the case (\ding{172}No) of the separation moment, we similarly generate $\mathcal{I}_\text{key}$ from the video segment after $\hat{T}_\text{s}$ to find a more obvious later separation. In the case (\ding{172}Yes$\rightarrow$\ding{173}No) of both the contact and separation moments, we generate $N_\text{key}$ frames with the lower hand velocities than the ones at $\hat{T}_\text{c}$ and $\hat{T}_\text{s}$ respectively. We anticipate that the new keyframes with lower hand velocities are more likely to encompass the correct contact/separation timestamp. According to the visual and dynamic cues, we ultimately generate more reasonable $\mathcal{I}_\text{key}$ in the second-round self-adaptive sampling, and construct a refined grid image $G$ for the following VLM-based reasoning. We therefore close the TIL loop and obtain more accurate contact and separation moment estimation $\hat{T}^{*}_\text{c}$ and $\hat{T}^{*}_\text{s}$.

As a reminder, EgoLoc first determines the contact timestamp and then identifies the separation timestamp from the subsequent segment. The two timestamps are not determined simultaneously, as the estimated contact timestamp helps narrow the range for the separation timestamp, enhancing its estimation accuracy. We also empirically found that a single round of closed-loop feedback significantly improves TIL performance, but additional rounds yield diminishing returns and substantially reduce inference efficiency. Therefore, we advocate using a single round of feedback in EgoLoc.

\begin{figure}[t]
  \centering
    \includegraphics[width=1\linewidth]{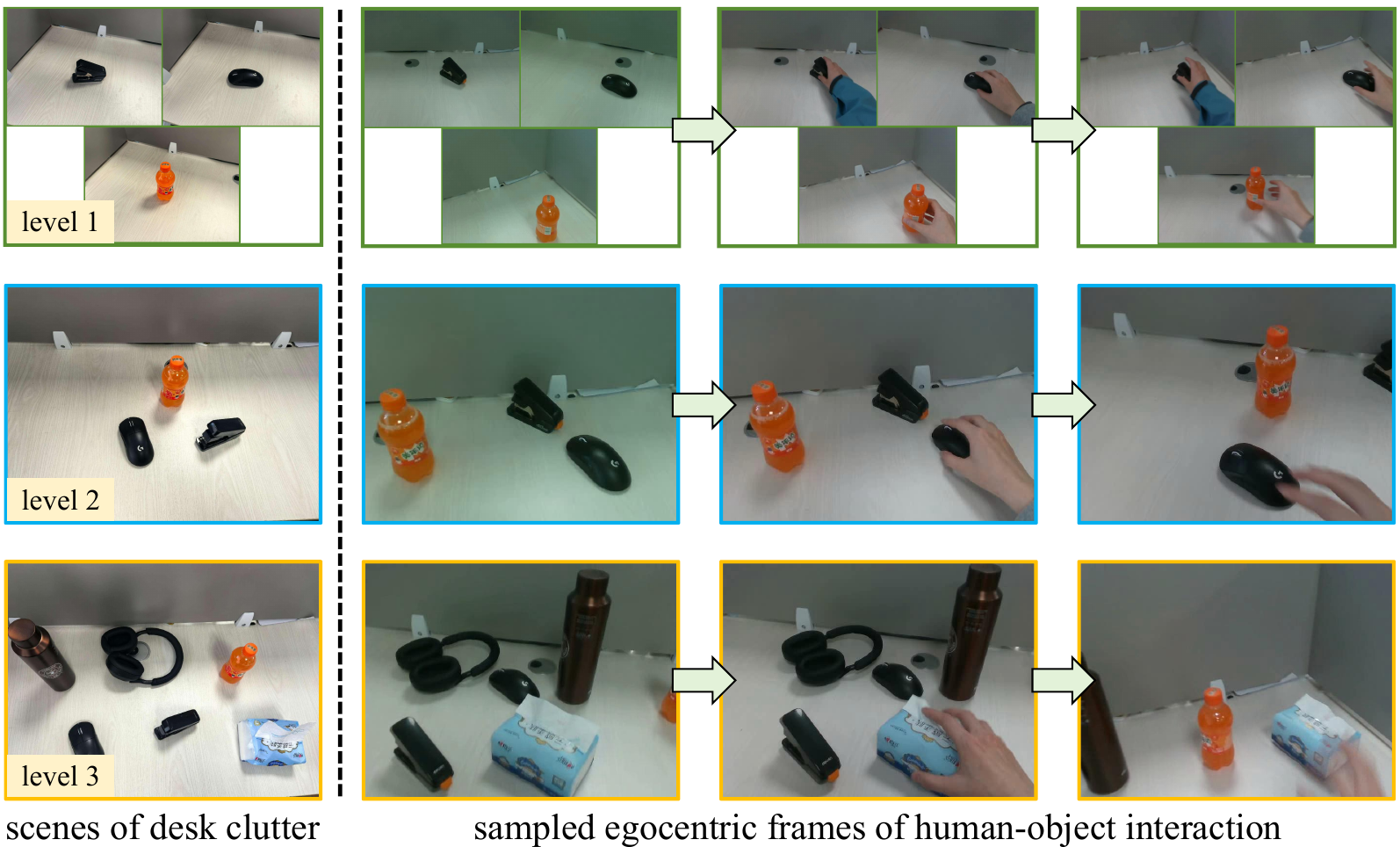}
  \caption{Visualization of the egocentric HOI data in our newly proposed DeskTIL benchmark.}
  \label{fig:benchmark}
  \vspace{-0.7cm}
\end{figure}

\begin{table*}[t]
\setlength{\tabcolsep}{17pt}
\center
\renewcommand\arraystretch{1}
\caption{Comparison of performance on temporal interaction localization on EgoPAT3D-DT and DeskTIL. Best and secondary results are viewed in \textbf{bold black} and \myblue{blue} colors respectively.}
\vspace{-0.15cm}
\begin{tabular}{l|cccccc}
\toprule
\multicolumn{1}{l|}{\multirow{2}{*}{Approach}}   & \multicolumn{6}{c}{EgoPAT3D-DT}  \\ \cmidrule{2-7} 
\multicolumn{1}{c|}{}                                                                               & SR($\gamma=1$)\,$\uparrow$    & SR($\gamma=2$)\,$\uparrow$ & SR($\gamma=3$)\,$\uparrow$   & MAE\,$\downarrow$ & MoF\,$\uparrow$   & IoU\,$\uparrow$   \\ \cmidrule{1-7}                 
Greedy VLM                 &0.149       &0.243       &0.317          &7.641    &0.533    &0.362\\
T-PIVOT ($N_\text{adj}=2$)   & 0.192    &0.340 	 &0.481   &4.654 	  &0.739   &0.693  \\ 
T-PIVOT ($N_\text{adj}=3$)   & 0.291     & 0.430	 & 0.551  & 3.867	  &0.780   & 0.727 \\ 
T-PIVOT ($N_\text{adj}=4$)   & 0.260     &0.437	 & 0.525  & 4.500	  &0.708    & 0.637 \\ 
Ours ($N_\text{adj}=2$) & \textbf{0.552}   & \textbf{0.734} 	& \textbf{0.857}  	& \textbf{1.864}	 & \textbf{0.907} 	& \textbf{0.868}\\ 
Ours ($N_\text{adj}=3$) & \myblue{0.378}    & \myblue{0.485} 	& \myblue{0.777}  	& \myblue{2.520}	 & \myblue{0.874} 	& \myblue{0.843}\\
Ours ($N_\text{adj}=4$) & 0.302   & 0.427 	& 0.693	& 3.323  & 0.818   & 0.781\\ \midrule
\multicolumn{1}{l|}{Approach}   & \multicolumn{6}{c}{DeskTIL}  \\ \midrule
Greedy VLM      &0.171       &0.285       &0.386        &8.608    &0.466    &0.252\\
T-PIVOT ($N_\text{adj}=2$)   &  0.535  &0.814 	 &0.896   & 1.767	    & 0.884 &0.756 \\
T-PIVOT ($N_\text{adj}=3$)   & 0.528     & 0.811	 & 0.908  & 1.803	 & 0.880  & 0.750
   \\ 
T-PIVOT ($N_\text{adj}=4$)   & 0.531     &0.808	 & 0.885  & 1.881	  &0.876    & 0.744 \\ 
Ours ($N_\text{adj}=2$) & \myblue{0.591}   & \myblue{0.818} 	& \myblue{0.909}  	& \myblue{1.591}	 & \myblue{0.890} 	& \myblue{0.809}
     \\ 
Ours ($N_\text{adj}=3$) & 0.541    & 0.796 	& 0.898 	& 1.827	 & 0.865	& 0.762
     \\ 
Ours ($N_\text{adj}=4$) & \textbf{0.655}    & \textbf{0.894} 	& \textbf{0.929}  	& \textbf{1.381}	 & \textbf{0.906} 	& \textbf{0.817}
     \\ 
\bottomrule
\end{tabular}
\label{tab:compare_main}
\vspace{-0.2cm}
\end{table*}

\begin{figure*}[t]
  \centering
  \includegraphics[width=1\linewidth]{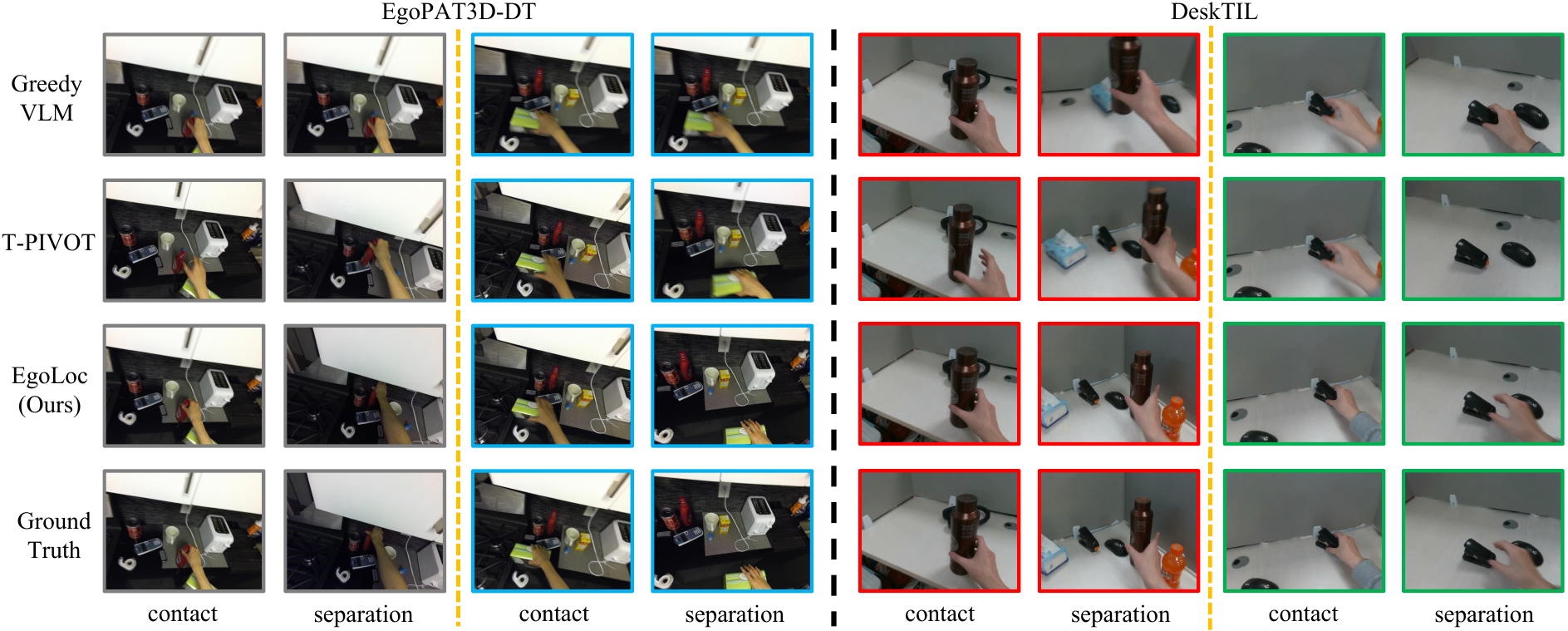}
  \caption{Visualization of the image frames at the estimated and ground-truth contact/separation timestamps for temporal interaction localization. Here we show the results achieved by each method under their optimal performance conditions in Tab.~\ref{tab:compare_main}.}
  \label{fig:viz_exp}
  \vspace{-0.6cm}
\end{figure*}

\section{Experiments}
\label{sec:exp}

\subsection{Datasets and Baselines}
\label{sec:setups}

We first utilize the EgoPAT3D dataset~\cite{li2022egocentric} to evaluate the TIL performance of our proposed EgoLoc. Specifically, we use the subset of the video clips processed by~\cite{bao2023uncertainty}, known as EgoPAT3D-DT, which contains more fine-grained HOI sequences to test TIL accuracy. 
We adopt their original frame rate 30\,FPS. 
Moreover, we build a new temporal interaction localization benchmark dubbed DeskTIL, considering different levels of desk clutter shown in Fig.~\ref{fig:benchmark}. We use a head-mounted RealSense D435i to collect 50 video sequences per level, with an average duration of around 6\,s per sequence. Each video in DeskTIL contains a single HOI process, and is downsampled to 5\,FPS for TIL evaluation under a low frame rate constraint. We manually annotate the contact and separation timestamps for each video in both datasets.

We select T-PIVOT~\cite{wake2024open} as our primary baseline, as it is the pioneering zero-shot method achieving SOTA TAL performance. We adapt it to TIL by providing the same universal text prompt as the one in our VLM-based reasoning module. It uniformly samples $N_\text{adj}$ frames to construct the grid image. We also design a new TIL baseline \textit{Greedy VLM}. Without any sampling operation, it directly uses the holistic image sequence to construct the grid image for each video, which is then fed to the VLM along with the universal text prompt to output contact and separation timestamps.

\vspace{-0.1cm}
\subsection{EgoLoc Configurations}
\label{sec:config}

In the self-adaptive sampling strategy, the frame
time interval $\delta$ is $1/30$\,s and $1/5$\,s for EgoPAT3D-DT and our DeskTIL benchmark respectively. The number of keyframes in $\mathcal{I}_\text{key}$ is set as $N_\text{key}=4$ for both datasets. We set $N_\text{adj}$ to $2\sim4$ to construct the grid image in the following experiments. In the VLM-based reasoning module, we exploit GPT-4o as the VLM-based localizer by default, and the universal text prompt is designed as \textit{Choose the number that is closest to the moment when `Grasping the object' started/ended}. As to the closed-loop feedback mechanism, GPT-4o is also used as the VLM-based discriminator, with the text prompt as \textit{Determine whether the hand and the object are in obvious contact rather than separation}. The velocity threshold $\mu_\text{vel}\%$ for the speed-based discriminator is $30\%$. 
For EgoLoc and all the other baselines, we perform 5 TIL trials for each video and report the averaged evaluation metrics across the 5 trials. 

\vspace{-0.1cm}
\subsection{Evaluation metrics}
\label{sec:eval_metrics}
We exploit the metrics MoF and IoU widely used in the TAL literature, to evaluate the performance of temporal interaction localization in this work. Moreover, considering the fine-grained nature of the TIL task, we introduce two new metrics, SR and MAE, to comprehensively quantify the accuracy of contact and separation moment estimations.

\noindent \textbf{MoF (Mean over Frames)} represents the percentage of the frames with correctly estimated stages (pre-contact, in-contact, or post-contact) out of the total frames for each video. We will report the average MoF across all videos.

\noindent  \textbf{IoU (Intersection over Union)} measures the overlap between the estimated and ground-truth segments of the in-contact stage for each video. It is calculated as the ratio of the intersection to the union of the two frame sets. We will report the average value of this metric across all videos.

\noindent  \textbf{SR (Success Rate)} calculates the proportion of successfully matched contact/separation timestamps between the estimations and ground-truth annotations 
considering all video clips. We set different tolerance ranges $\gamma$ for this metric. One estimation can be regarded as a success if the frame interval between the estimated and ground-truth timestamps falls within the preset tolerance range.

\noindent  \textbf{MAE (Mean Absolute Error)} measures the absolute error, i.e., the frame interval between the estimated and ground-truth timestamps for each video. We will report the average value of this metric across all videos.

\vspace{-0.1cm}
\subsection{Main Results}
\label{sec:comparison_with_prior}

We first compare the TIL performance of our proposed EgoLoc method with the selected baselines. 
Tab.~\ref{tab:compare_main} and Fig.~\ref{fig:viz_exp} show that, with the devised self-adaptive sampling strategy and closed-loop feedback mechanism, our EgoLoc outperforms the baselines on all evaluation metrics for both the EgoPAT3D-DT dataset and our DeskTIL benchmark. EgoLoc achieves the best performance at $N_\text{adj}=2$ for EgoPAT3D-DT and $N_\text{adj}\!=\!4$ for DeskTIL. This indicates that we need more frames to construct the grid image as the visual prompt for the one-step inference of the VLM-based localizer on DeskTIL than EgoPAT3D-DT. The reason could be that in the cluttered DeskTIL scenes, the 2D hand motion often traverses the pixel regions of multiple objects. This requires more frames in the grid image to accurately capture the state transition of the actual target object during interaction. We also compare the inference speed of EgoLoc and T-PIVOT both with the same $N_\text{adj}=4$. T-PIVOT requires approximately 1.3/1.9 times the inference time of EgoLoc for each video in EgoPAT3D-DT/DeskTIL. EgoLoc implements temporal localization more efficiently thanks to the high-quality initial guesses from 3D hand motion analysis.

\begin{table}[t]
\setlength{\tabcolsep}{5pt}
\center
\renewcommand\arraystretch{0.6}
\caption{Ablation study on 3D perception with EgoPAT3D-DT. Best results are viewed in \textbf{bold black}.}
\vspace{-0.15cm}
\begin{tabular}{l|cc|cc}
\toprule
\multicolumn{1}{l|}{\multirow{2}{*}{Approach}}   & \multicolumn{4}{c}{EgoPAT3D-DT}    \\ \cmidrule{2-5} 
\multicolumn{1}{c|}{}                                                                               & SRc($\gamma=3$)\,$\uparrow$    & MAE\,$\downarrow$ & MoF\,$\uparrow$    & IoU\,$\uparrow$    \\ \cmidrule{1-5}    
EgoLoc-2D ($N_\text{adj}=2$) & 0.795	& 2.577  & 0.889   & 0.848\\
EgoLoc-2D ($N_\text{adj}=3$) & 0.715	& 2.714  & 0.856   & 0.837\\
EgoLoc-2D ($N_\text{adj}=4$) & 0.621	& 3.705  & 0.813   & 0.778\\
\midrule
EgoLoc ($N_\text{adj}=2$) & \textbf{0.857}	& \textbf{1.864}  & \textbf{0.907}   & \textbf{0.868}\\
EgoLoc ($N_\text{adj}=3$) & 0.777	& 2.520  & 0.874   & 0.843\\
EgoLoc ($N_\text{adj}=4$) & 0.693	& 3.323  & 0.818   & 0.781\\ \bottomrule
\end{tabular}
\label{tab:abla_velocity}
\vspace{-0.6cm}
\end{table}

\subsection{Ablation Studies}
\label{sec:exp_albation}
\textbf{3D perception}. 
We first evaluate the impact of incorporating 3D perception data (depth images) to extract hand velocities on EgoLoc performance. Specifically, we conduct a baseline EgoLoc-2D that only uses 2D RGB images as input. It implements 2D hand velocity extraction in the self-adaptive sampling strategy, and performs 2D speed-based discrimination in the closed-loop feedback mechanism both in the image plane of the first frame.
The experimental results in Tab.~\ref{tab:abla_velocity} show that incorporating 3D perception information significantly improves the baseline relying solely on 2D observations on all the evaluation metrics. This indicates that exploiting 3D hand velocities from 3D perception enhances the model's spatial awareness of actual human motion characteristics, avoiding depth ambiguity and scale aliasing of 2D observations. It also demonstrates that the 3D hand velocity-based sampling produces high-quality keyframes for VLM reasoning, leading to better TIL performance.

\begin{figure}[t]
  \centering
    \includegraphics[width=0.95\linewidth]{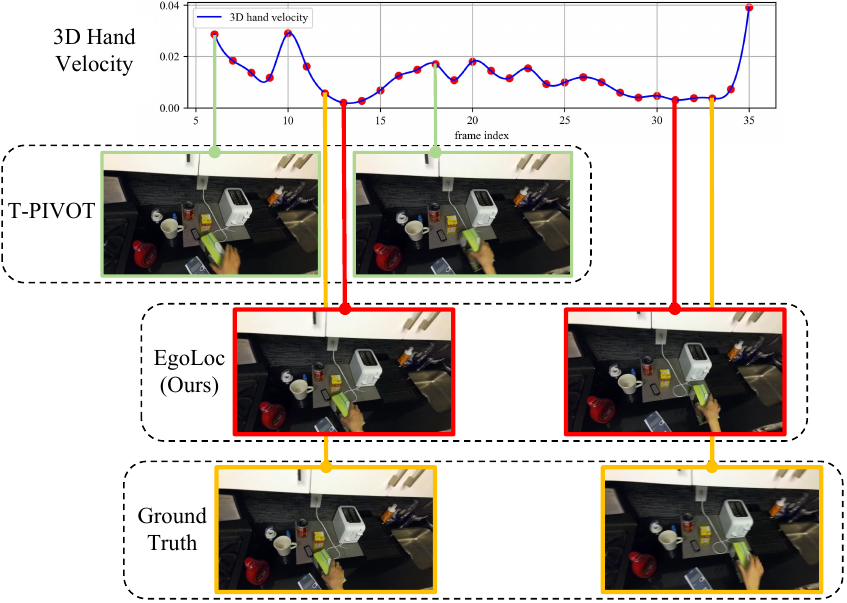}
  \caption{Visualization of the ground-truth contact/separation frames and the anchor frames sampled by T-PIVOT and our EgoLoc.}
  \label{fig:keyframes}
  \vspace{-0.3cm}
\end{figure}

As mentioned in Sec.~\ref{sec:sas}, the anchor frame $I_\text{init}$ is sampled as the center image of the grid image $G$ for our VLM-based localizer. EgoLoc discards the initial uniform sampling in T-PIVOT\cite{wake2024open} at the beginning of VLM reasoning. In Fig.~\ref{fig:keyframes}, we compare the anchor frame $I_\text{init}$ sampled in our self-adaptive sampling strategy and the counterpart from the initial uniform sampling by T-PIVOT. EgoLoc samples more reasonable anchor frames closer to the ground-truth contact/separation timestamps than T-PIVOT.

\begin{table}[t]
\setlength{\tabcolsep}{4.3pt}
\center
\caption{Ablation study on the closed-loop feedback mechanism with EgoPAT3D-DT. Best results are viewed in \textbf{bold black}.}
\vspace{-0.1cm}
\renewcommand\arraystretch{0.6}
\begin{tabular}{cc|cc|cc}
\toprule
\multicolumn{2}{c|}{Discriminator}  &\multicolumn{4}{c}{EgoPAT3D-DT}   \\ \midrule
VLM-based & Speed-based  & SR ($\gamma=3$)\,$\uparrow$    & MAE\,$\downarrow$ & MoF\,$\uparrow$    & IoU\,$\uparrow$   \\ \midrule          
\ding{55} &  \ding{55}   &0.716 &2.520 &0.889  &0.843 \\
\ding{55} & \ding{51}   &0.827 &2.224 &0.885  &0.835
  \\
\ding{51} &\ding{55}  &0.763 &2.340 &0.891 &0.846  \\
\ding{51}  & \ding{51} & \textbf{0.857} 	& \textbf{1.864}	 & \textbf{0.907}	  & \textbf{0.868}\\ \bottomrule
\end{tabular}
\label{tab:ala_on_feedback1}
\vspace{-0.5cm}
\end{table}

\begin{table}[t]
\setlength{\tabcolsep}{8.5pt}
\center
\renewcommand\arraystretch{0.6}
\caption{Ablation study on VLMs with DeskTIL. Best results are viewed in \textbf{bold black}.}
\vspace{-0.1cm}
\begin{tabular}{l|cc|cc}
\toprule
\multicolumn{1}{l|}{\multirow{2}{*}{Approach}}   & \multicolumn{4}{c}{DeskTIL}    \\ \cmidrule{2-5} 
\multicolumn{1}{c|}{}                                                                               & SR ($\gamma=3$)\,$\uparrow$    & MAE\,$\downarrow$ & MoF\,$\uparrow$    & IoU\,$\uparrow$    \\ \cmidrule{1-5}    
GPT-4o mini  & 0.654	& 3.442  & 0.779   & 0.630\\
GPT-4 Turbo  & 0.803	& 2.364  & 0.823   & 0.658\\
GPT-4o  & \textbf{0.929}	&\textbf{1.381}   &\textbf{0.906}    &\textbf{0.817}  \\
\bottomrule
\end{tabular}
\label{tab:aba_vlm}
\vspace{-0.2cm}
\end{table}

\textbf{Closed-loop feedback mechanism}.
Next, we ablate our proposed closed-loop feedback mechanism of EgoLoc ($N_\text{adj}=2$). We present the TIL performance of the baselines including EgoLoc without closed-loop feedback, with only the VLM-based discriminator using visual cues, with only the speed-based discriminator using dynamic cues, and the vanilla EgoLoc with both discriminators. 
As in Tab.~\ref{tab:ala_on_feedback1}, either VLM-based or speed-based discriminators individually improve TIL performance compared to the baseline without closed-loop feedback. In addition, the vanilla EgoLoc with both discriminators achieves the best performance. This validates that the combination of the VLM-based and speed-based discriminators comprehensively helps to assess the first-round TIL estimations with both visual and dynamic cues, resulting in more accurate second-round estimations.

\textbf{VLM selection}. We also present EgoLoc ($N_\text{adj}=4$) performance with different VLMs on the DeskTIL benchmark, including GPT-4o, GPT-4o mini, and GPT-4 Turbo. Tab.~\ref{tab:aba_vlm} shows that GPT-4o generates the best TIL estimations, which is the default VLM used in our proposed EgoLoc.

\vspace{-0.2cm}
\subsection{Study on Inference Uncertainties}
Open-loop LLM/VLM reasoning has inherent inference uncertainties due to the nature of probabilistic generative models~\cite{ye2025benchmarking}, which harms the system stability of interaction localization. To demonstrate that our closed-loop TIL paradigm holds low estimation uncertainties for temporal interaction localization, we further present the standard deviation of each evaluation metric across 5 trials in Tab.~\ref{tab:uncertainty}. As seen, our EgoLoc generates more stable estimation results since it has the lowest standard deviations for all these metrics. Notably, the proposed closed-loop feedback can effectively reduce inference uncertainties of EgoLoc thanks to the second-round refinement.

\begin{table}[t]
\setlength{\tabcolsep}{5pt}
\center
\renewcommand\arraystretch{0.6}
\caption{Ablation study on inference uncertainties with EgoPAT3D-DT. Best results are viewed in \textbf{bold black}.}
\vspace{-0.1cm}
\begin{tabular}{l|cc|cc}
\toprule
\multicolumn{1}{l|}{\multirow{2}{*}{Approach}}   & \multicolumn{4}{c}{EgoPAT3D-DT}    \\ \cmidrule{2-5} 
\multicolumn{1}{c|}{}                                                                               & SR$^{\dag}$($\gamma=3$)\,$\downarrow$    & MAE$^{\dag}$\,$\downarrow$ & MoF$^{\dag}$\,$\downarrow$    & IoU$^{\dag}$\,$\downarrow$    \\ \cmidrule{1-5}    
Greedy VLM  & 0.026	& 0.240  & 0.029   & 0.037\\
T-PIVOT  & 0.066	& 0.839  & 0.078   & 0.081\\
Ours (w/o feedback)  & 0.023	& 0.187  & 0.022   & 0.029 \\
Ours (w/ feedback)  & \textbf{0.013}	& \textbf{0.089}  & \textbf{0.006}   & \textbf{0.011} \\
\bottomrule
\end{tabular}
\vspace{-0.1cm}
\begin{flushleft}
$\dag$: standard deviation of the metric.
\end{flushleft}
\vspace{-0.7cm}
\label{tab:uncertainty}
\end{table}

\vspace{-0.1cm}
\section{Conclusion}
\vspace{-0.1cm}
\label{sec:discuss}
In this paper, we propose a novel temporal interaction localization method EgoLoc, which can effectively find the timestamps at which a human hand makes contact with and separates from a target object in egocentric videos. We design a self-adaptive sampling strategy to generate high-quality initial guesses for VLM-based reasoning, and develop a closed-loop feedback mechanism to further refine the TIL results. Compared to the existing TAL paradigms, EgoLoc identifies more fine-grained human-object interaction timings, generalizes well without pertaining, better perceives interaction in 3D observations, and finally closes the TIL reasoning loop. The experimental results on the publicly available dataset and our newly proposed benchmark demonstrate the superiority of EgoLoc, and the rationality of its components in the task of egocentric temporal interaction localization. In future work, we will eliminate the assumption in EgoLoc that contact timestamps are located in the first half of the video clip through pre-executed TAL, and test EgoLoc on longer untrimmed videos.

\bibliographystyle{ieeetr}

\footnotesize{
\bibliography{root}}

\end{document}